\documentclass[runningheads]{llncs}

\usepackage{eccv}

\usepackage{eccvabbrv}

\usepackage{graphicx}
\usepackage{booktabs}

\usepackage[accsupp]{axessibility}  

\usepackage{hyperref}

\usepackage{orcidlink}

\usepackage{multirow}
\usepackage{gensymb}
\usepackage{tabularx}
\usepackage{booktabs} 
\usepackage{xcolor} 

\begin{document}

\title{NocPlace: Nocturnal Visual Place Recognition via Generative and Inherited Knowledge Transfer} 

\titlerunning{NocPlace: Nocturnal Visual Place Recognition}

\author{Bingxi Liu\inst{1, 2} \and
Yiqun Wang\inst{3} \and
Huaqi Tao\inst{1} \and
Tingjun Huang\inst{1} \and \\
Fulin Tang\inst{4} \and
Jingqiang Cui\inst{2} \and
Yihong Wu\inst{4} \and
Hong Zhang\inst{1} }

\authorrunning{B.Liu et al.}

\institute{
Southern University of Science and Technology, Shenzhen, China \and
Peng Cheng Laboratory, Shenzhen, China \and
Chongqing University, Chongqing, China \and
Institute of Automation, Chinese Academy of Sciences, Beijing, China}

\maketitle

\begin{abstract}
  Visual Place Recognition (VPR) is crucial in computer vision, aiming to retrieve database images similar to a query image from an extensive collection of known images. However, like many vision tasks, VPR always degrades at night due to the scarcity of nighttime images. Moreover, VPR needs to address the cross-domain problem of night-to-day rather than just the issue of a single nighttime domain. In response to these issues, we present NocPlace, which leverages generative and inherited knowledge transfer to embed resilience against dazzling lights and extreme darkness in the global descriptor. First, we establish a day-night urban scene dataset called NightCities, capturing diverse lighting variations and dark scenarios across 60 cities globally. Then, an image generation network is trained on this dataset and processes a large-scale VPR dataset, obtaining its nighttime version. Finally, VPR models are fine-tuned using descriptors inherited from themselves and night-style images, which builds explicit cross-domain contrastive relationships. Comprehensive experiments on various datasets demonstrate our contributions and the superiority of NocPlace. Without adding any real-time computing resources, NocPlace improves the performance of Eigenplaces by 7.6\% on Tokyo 24/7 Night and 16.8\% on SVOX Night.
  \keywords{Visual place recognition \and Domain adaption \and Transfer learning}
\end{abstract}

\section{Introduction}
\label{sec:introduction}

\begin{figure}[t] 
\center
{\includegraphics[width=0.95\textwidth] {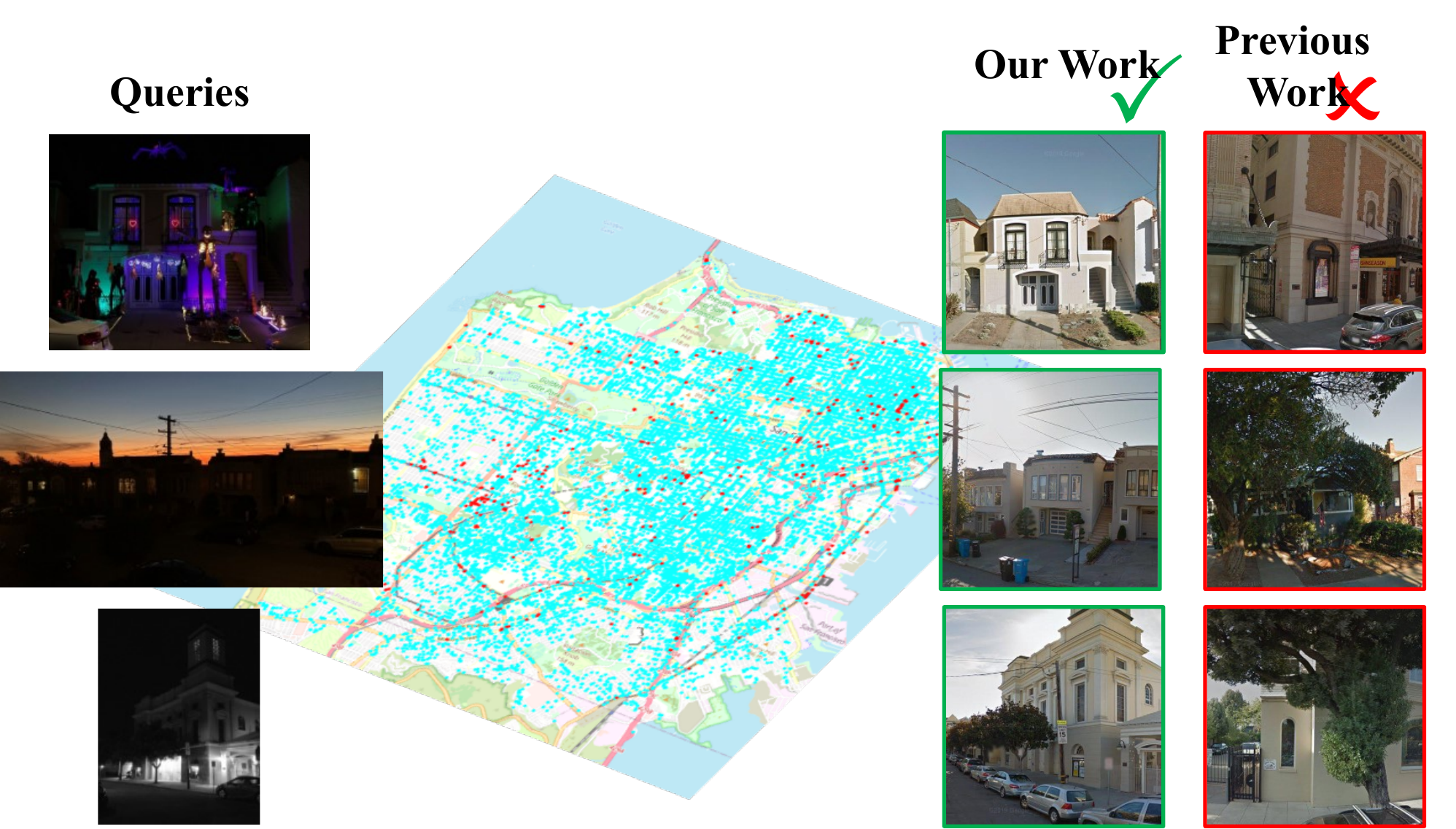}}
\caption{\textbf{A demo for NocPlace.} The center of this figure presents a map of San Francisco. 2.8 million daytime database images and 466 nighttime query images from SF-XL \cite{cosplace} are projected onto this map via GPS tags, corresponding to the dense cyan-blue and sparse red dots, respectively. Despite the dazzling lights and extreme darkness, our method achieves higher recall@N than previous methods.} 
\label{fig:hightlight}
\end{figure}

Visual Place Recognition (VPR) is a foundational task in computer vision \cite{netvlad, patch_netvlad, 3d_model, 247_dataset, repetitive_structures, vpr_bench} and intelligent robotics \cite{kitti, tro_survey, robotcar, seqslam, appsvr, msls, lu2023aanet, wu2019deephashing, tutorial, keetha2023anyloc} and is mainly used in Augmented Reality (AR) \cite{hloc, feigl2020localization} autonomous systems \cite{robotcar, huawei}. VPR aims to retrieve similar database images to a given query image by comparing it against a vast set of known database images. Existing literature identifies several challenges for VPR, including database scale \cite{cosplace}, viewpoint shifts \cite{viewpoint, eigenplaces}, repeated structures \cite{repetitive_structures}, structural modifications \cite{change}, occlusions \cite{netvlad}, visual scale differences \cite{scalenet}, illumination changes \cite{todaygan, delg, robotcar, tro_survey, lamar, long_vl, 247_dataset, msls, vpr_bench, schubert2021makes, ie_vpr, lengyel2021zero, dsa, neubert2022seer}, and seasonal transitions \cite{long_vl, seqslam, nordland}. Recent VPR approaches primarily leverage large-scale datasets \cite{cosplace, st_lucia, GLDv1, 247_dataset, gldv2, gsv} and utilize a neural network to map images into an embedding space, thereby effectively distinguishing images from different locations.

However, night-to-day VPR is constrained by the current training datasets. On the one hand, recent studies \cite{netvlad, ali2023mixvpr, eigenplaces, gsv_cities, cosplace} leveraging Google Street View (GSV)\footnote{\url{https://developers.google.com/streetview/}} have produced large-scale VPR datasets covering nearly all challenges \textit{except} those in nighttime conditions. On the other hand, while some datasets specifically targeting night-to-day VPR datasets \cite{robotcar, seqslam} have been proposed and shown the benefit of incorporating nighttime data for cross-domain recognition \cite{neubert2022seer, dsa}, their scale or data acquisition methods impede the training of a \textit{universally} night-to-day VPR model. What is evident is that building nighttime datasets that \textit{align} with commercial products like GSV \cite{gsv} is challenging.

Due to the above restrictions, image enhancement algorithms (IE) \cite{sien, dual, lime, zero, ie_vpr} and night-to-day Image-to-Image (I2I) translation \cite{todaygan} may be a potential approach. These methods, however, present several limitations: i) Introducing an auxiliary model increases computational demands. ii) Most low-light enhancement models rely on images with diverse exposure rather than real night-to-day transitions. Some images do not even come from the outdoors \cite{lol_dataset}. iii) Imperfect night-to-day translations might adversely affect VPR performance \cite{todaygan}.

In this paper, we reverse the night-to-day thinking \cite{todaygan} and study how to transfer night knowledge directly into neural networks. Further, we study the use of daytime knowledge to guide the training of nighttime models. As shown in \cref{fig:hightlight}, our method can improve the performance of nighttime VPR without consuming additional real-time computing resources.

Our contributions can be summarized as follows:

\begin{itemize}

\item We propose NocPlace, a scalable Nocturnal Visual Place Recognition method that significantly narrows the VPR performance gap between daytime and nighttime domains.

\item We propose NightCities, an unpaired day-night urban scene image dataset. We utilize this dataset and an I2I translation network to transfer night knowledge to a large-scale daytime VPR dataset.

\item We propose the inherited knowledge transfer, which employs descriptors inherited from previous VPR models to guide the training of new VPR models.

\item We propose a partial-divide-and-conquer-based retrieval strategy for practical applications, which balances storage and performance.

\item Extensive experiments demonstrate that NocPlace substantially outperforms previous methods. Specifically, our VGG16-based model excels over state-of-the-art ResNet50-based models on certain test sets.

\item The source code and trained models for NocPlace are available at \url{https://github.com/BinuxLiu/NocPlace}.

\end{itemize}

\section{Related Work}
\label{sec:related_work}

\noindent \textbf{Visual Place Recognition} can be categorized along three pivotal axes: model architecture, loss function, and dataset. About model architecture, Convolutional Neural Networks (CNNs) \cite{netvlad, schubert2020unsupervised} and Visual Transformers (ViTs) \cite{vaswani2017attention, transvpr, dvg} emerge as the primary choices for feature extraction. These are often followed by pooling or aggregation mechanisms like NetVLAD \cite{netvlad} and GeM \cite{dir}. In loss functions, triplet and classification loss are frequently employed to optimize Euclidean margins for enhanced feature embeddings \cite{dvg, netvlad}. Ge \etal~\cite{sfrs} utilized triplet loss and network output scores as self-supervised indicators for iterative training. However, triplet loss encounters challenges with hard negative sample mining. In response, Berton \etal~\cite{cosplace, eigenplaces} incorporated Large Margin Cosine (LMC) Loss \cite{cosface} in VPR, showcasing its superiority over triplet loss. On the dataset front, we observe two primary classifications: self-built datasets and web-sourced ones. The former, such as \cite{kitti, robotcar, seqslam}, often exhibit small scale, fixed frontal camera view, and little nighttime data. The latter can be sub-categorized into datasets like Google Landmark \cite{dir}, derived from Google Images, and those like SF-XL \cite{cosplace} and Pittsburgh \cite{netvlad}, sourced from GSV \cite{netvlad, gsv_cities, gsv, cosplace}. It is worth noting that the latter contains almost no night images. 

\noindent \textbf{Night Computer Vision} encompasses downstream tasks using nighttime images, which are divided into one-stage and two-stage approaches. The one-stage methods utilize nighttime data for training \cite{risp, vpr_bench}. Cui \etal~\cite{risp} developed a technique for reverse ISP and dark processing, leading to an end-to-end model for object detection in dark conditions. Additionally, Xu \etal~\cite{gan_ss} presented a GAN-centric strategy for semantic segmentation using transformed nighttime imagery. Attila \etal~\cite{lengyel2021zero} introduced the color invariant layer, reducing the day-night distribution shift in feature map activations. In contrast, two-stage approaches \cite{todaygan, gan_ss} first transform nighttime images into daytime-style images before executing the downstream tasks. Anoosheh \etal~\cite{todaygan} introduced a CycleGAN-based method to convert nighttime images for retrieval-based localization. Neubert and Schubert \cite{neubert2022seer} proposed a simple algorithm to learn domain-specific descriptors from a small database. However, the two-stage methods are criticized for their limited generalizability \cite{todaygan} and heightened computational demands. Meanwhile, the one-stage approach faces challenges due to the lack of extensive training datasets. In response, we propose a pioneering framework by converting daytime images to nighttime versions for enhanced VPR training.

\noindent \textbf{Image-to-Image Translation} pertains to the intersection of vision and graphics, aiming to establish a mapping between an input and an output image \cite{unpaired}. These methods are primarily categorized based on the nature of their training sets: pixel-level matched (paired) I2I \cite{pix2pix} or unpaired I2I \cite{unpaired, dualgan}. Collecting extensive paired day-night images in real-world scenarios is notably challenging. Hence, our focus is on the latter training set type. Two predominant strategies for unpaired I2I translation exist bi-directional mapping and uni-directional mapping. Bi-directional mapping, encompassing CycleGAN \cite{unpaired} and DualGAN \cite{dualgan}, is based on the cycle-consistency constraint, ensuring the transposed image can undergo reverse mapping for reconstruction. The bi-directional mapping can be overly restrictive, prompting the evolution of uni-directional mapping methods. For instance, DistanceGAN \cite{distance} conserves inter-class distances within domains, while GCGAN \cite{gcgan} upholds geometry consistency between inputs and outputs. The CUT \cite{cut} amplifies mutual information between inputs and outputs via contrastive loss. The diffusion model \cite{egsde} recently has shown exceptional performance. However, we found that it consumes considerable computational resources and is unsuitable for generating large-scale data. After thoroughly comparing different methods, we use NEG-CUT \cite{negcut} for nighttime data generation.

\noindent \textbf{Knowledge Transfer} can be categorized in various ways based on the problems it addresses, such as domain adaptation \cite{domain_adaptation}, multi-task learning \cite{multi_task}, meta-learning \cite{meta_learning}, and more. This paper focuses on domain adaptation, aiming to adapt models trained in one domain (source) to perform optimally in another domain (target) where the data distribution might differ. Previous Night-to-Day VPR studies primarily focused on labeled target domains, a relatively straightforward problem \cite{wu2019deephashing, ye2023condition, qin2021structure}. Cao \etal~\cite{qin2021structure} incorporated Canny edge descriptors and the Wasserstein distance measure for domain adaptation in VPR tasks. When dealing with labeled source and unlabeled target domains, prior research can be broadly divided into two categories: i) generation of target domain data, as exemplified by SimGAN \cite{simgan}, which transfers samples from the source to the target domain while preserving consistent labels. ii) adversarial learning of domain-invariant features. Ganin \etal~introduced DANN \cite{ganin2016domain}, which learns domain-invariant feature distributions by minimizing the source domain classification loss and maximizing the domain classification loss. We present the generative knowledge transfer, which aligns with the first category, and introduce inherited knowledge transfer that minimizes the discrepancy between domain feature distributions.

\begin{figure}[t] 
\center
{\includegraphics[width=\textwidth] {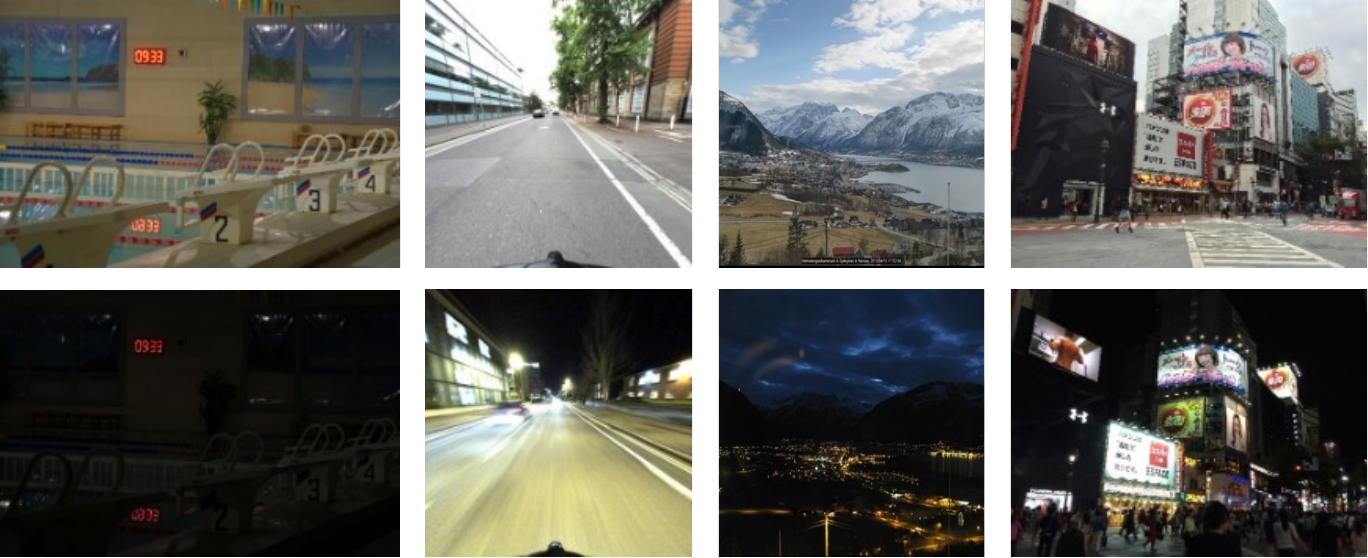}}
\caption{ \textbf{Potential Datasets for Day-to-Night Translation.} From left to right, they represent (a) LOL \cite{lol_dataset}, (b) RobotCar \cite{robotcar}, (c) day2night \cite{day2night}, and (d) Tokyo 24/7 \cite{247_dataset}.} 
\label{fig:previous_datasets}
\end{figure}

\section{Day-to-Night Translation Datasets}
\label{sec:day_to_night_translation_datasets}

This section introduces the motivation behind constructing the NightStreets and NightCities datasets. Subsequently, we delve into details regarding the composition of these datasets. It is important to note that these datasets are prerequisites for our methodology and evaluation.

\noindent \textbf{Motivation for Datasets Construction.} We present an observation: existing large-scale, multi-view VPR training sets, ranging from 30k images \cite{netvlad} to a remarkable 41.2M images \cite{cosplace}, mainly consist of daytime images and lack nighttime images. This omission significantly affects the representation capabilities of previous work under nocturnal conditions, making them less optimal than in daytime scenarios. While some day-night VPR training sets \cite{robotcar, msls} have demonstrated the benefits of nighttime training data for improved cross-domain recognition, they suffer from limited generalization. Two main factors contribute to this limitation: i) the presence of only a single nighttime style in some datasets \cite{robotcar} and ii) a considerably smaller scale than daytime datasets. We also recognize two significant challenges: i) Constructing a nighttime VPR dataset that aligns with the GPS tags or 6 DoF poses of existing VPR datasets is exceptionally difficult. ii) Building a Day-Night VPR dataset from scratch, which matches the scale of current datasets, is also a daunting task. Given these challenges, we propose an indirect approach: utilizing I2I translation techniques to convert daytime images into a nighttime style. Consequently, there is an urgent need for datasets encompassing both day and night scenarios for training an I2I translation network.

Regrettably, our search for a suitable day-to-night I2I translation dataset was unsuccessful. As illustrated in \cref{fig:previous_datasets} (a), most low-light enhancement datasets are primarily concerned with low-light conditions \cite{lol_dataset, sien}, such as those induced by backlighting or weak exposure, rather than capturing the diverse nuances of real nighttime scenes. \cref{fig:previous_datasets} (b) showcases examples from the aforementioned day-night VPR dataset \cite{robotcar}. Models trained on these datasets have limited generalization capabilities for night-to-day VPR, suggesting that theoretically generated day-to-night scene data might also be inadequate. Furthermore, we observe recurring limitations in autonomous driving datasets, such as image blur and single camera orientation. \cref{fig:previous_datasets} (c) references several time-lapse photography datasets proposed in the previous I2I transfer research \cite{day2night, pix2pix}. However, these datasets tend to focus on distant views and skylines, deviating significantly from urban scenes, and a day-to-night time-lapse video contains higher redundancy than a few photos. As shown in \cref{fig:previous_datasets} (d), the night test sets of VPR can be used to train I2I, but it lacks fairness because it may lead to implicit data leakage.

\begin{figure}[tbp]
\center
{\includegraphics[width=\textwidth] {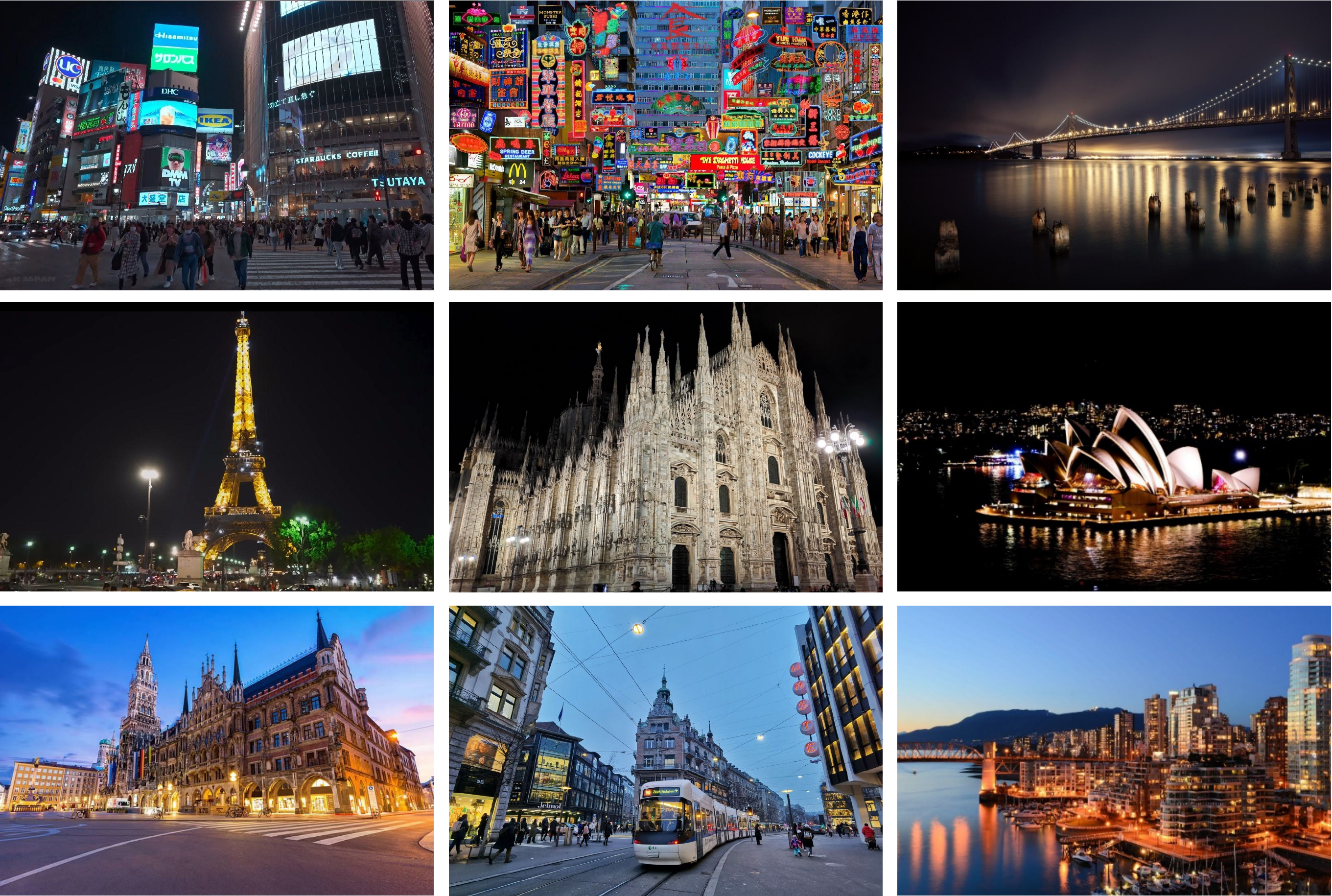}}
\caption{ \textbf{Examples of the NightCities dataset.} Only nighttime images are displayed here, and daytime images are shown in the supplementary material. The images are arranged from left to right and top to bottom, representing the following cities in order: Tokyo, Hong Kong, San Francisco, Paris, Milan, Sydney, Munich, Zurich, and Vancouver.}
\label{fig:nightcities_dataset}
\end{figure}

\noindent \textbf{The NightCities Dataset.} Consequently, we collected a large-scale unpaired day-night urban scene dataset named NightCities, as illustrated in \cref{fig:nightcities_dataset}. It encompasses 5,000 images from 60 cities globally, encapsulating diverse lighting styles, cultural vibes, and architectural features. Notably, the quantity of images aligns with representative datasets in the I2I translation domain \cite{choi2020stargan, gans, egsde}. The only annotation required was the categorization into day and night. We trained an unpaired I2I translation network on NightCities and used this network to process a subset of SF-XL, resulting in SF-XL-NC. It consists of a training set of 59,650 night-style images, a validation set of 7,993 night-style queries, and 8,015 database images.

\noindent \textbf{The NightStreets Dataset.} We also built a test set for evaluating I2I translation by reorganizing query images from Tokyo 24/7 v3 \cite{247_dataset} and Aachen Day/Night v1.1 \cite{vl_benchmarking}, as illustrated in \cref{fig:previous_datasets} (d). Tokyo 24/7 v3 \cite{247_dataset} comprises 375 daytime and nighttime images each, while Aachen Day/Night v1.1 \cite{vl_benchmarking} contains 234 daytime and 196 nighttime images. To distinguish it from the NightCities dataset, we call it NightStreets.

\section{Methodology}
\label{sec:methodology}

\begin{figure*}[t] 
\begin{center}
{\includegraphics[width=\textwidth] {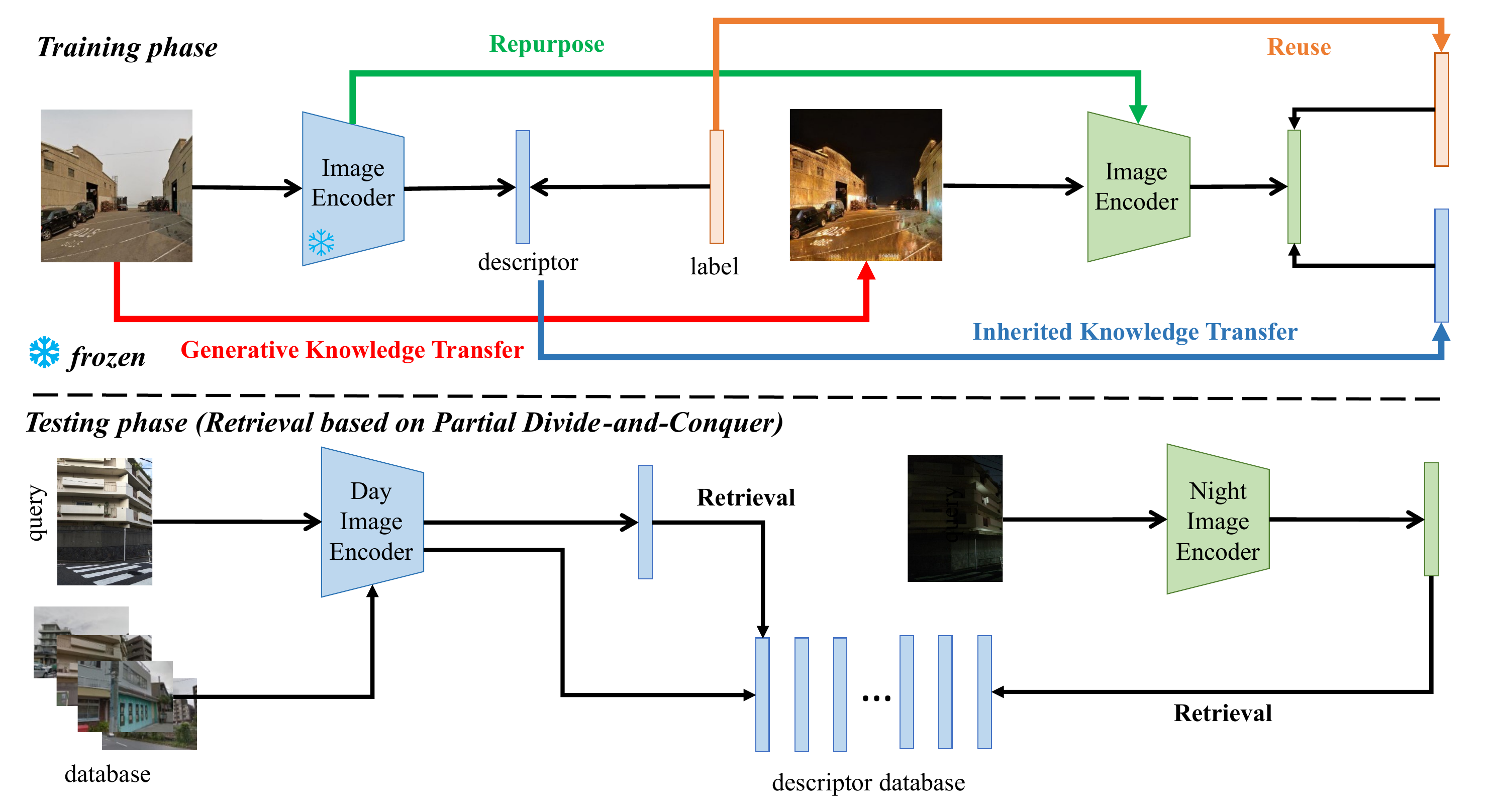}}
\end{center}
   \caption{\textbf{Schematic Diagrams of NocPlace.} In the training phase, we transfer the night knowledge into the training data and use the existing label information. We inherit the descriptors extracted from the corresponding data by the pre-trained daytime VPR model and use them to supervise the NocPlace model. During the test phase, NocPlace extracts features from the query images and queries them in the descriptor database extracted by the daytime VPR model.}
\label{fig:framework}
\end{figure*}

In this section, we detail the training and testing strategies of NocPlace, which encompass generative knowledge transfer, inherited knowledge transfer, and a divide-and-conquer retrieval approach. The upper part of \cref{fig:framework} corresponds to \cref{sec:gkt} and \cref{sec:ikt}, and the lower part corresponds to \cref{sec:pdc}.

\subsection{Generative Knowledge Transfer}
\label{sec:gkt}

Generative knowledge transfer aims to generate pseudo target domain data using labeled source domain data, especially in scenarios where labeled data in the target domain is hard to obtain.
The labels from the source domain data are then directly transferred to the corresponding generated data. Specifically, our objective is to maintain the scene's content from daytime images while coupling it with a nighttime style. For implementation, we employed NEG-CUT \cite{negcut}, an unpaired I2I generation method based on contrastive learning \cite{cut}. It maintains local content consistency by mining challenging image patches and learns the target domain style using a discriminator and adversarial loss \cite{pix2pix}.

Employing the VPR training framework \cite{cosplace, eigenplaces} with Large Margin Cosine (LMC) loss \cite{cosface}, we can either fine-tune the model using both source and generated domain data or train exclusively with the generated domain data. The former approach balances performance between daytime and nighttime, while the latter emphasizes nighttime performance.

\subsection{Inherited Knowledge Transfer}
\label{sec:ikt}

The advantage of LMC loss over triplet loss is that it avoids mining difficult samples, which makes it easier to achieve the current state-of-the-art results on large-scale training data sets \cite{cosplace, eigenplaces}. However, LMC loss also has the disadvantage that it cannot explicitly construct cross-domain contrastive relationships. Meanwhile, we assume a pre-trained VPR model has learned a representation suitable for daytime VPR. Therefore, we aim to minimize the distribution difference between the day and night domains to inherit the representation ability of the pre-trained model. Specifically, we save the global descriptors extracted from daytime images by the pre-trained model, align them with the nighttime images, and minimize the network outputs by Kullback–Leibler (KL) divergence.

We call the above method Inherited Knowledge Transfer (IKT). It is very similar to Knowledge Distillation (KD) \cite{hinton2015distilling} at a macro level but is significantly different in specific definition and implementation. KD involves using the outputs from a larger model (teacher) to guide the training of a smaller model (student) on the same dataset, aiming to enhance the student's performance. In most cases, the performance of this student model will not exceed that of the teacher model. In our case, we guide a new model with an older model of the \textit{same} architecture on data containing \textit{new} knowledge, aiming to \textit{enhance} the new model's performance.

In terms of implementation, IKT loss is also different from KD loss. IKT Loss is aligned with LMC loss and is calculated in cosine space. The formula for LMC loss is as follows:

\begin{equation}
 L_{\mathrm{LMC}} = \frac{1}{N}\sum_{i}{-\log{\frac{e^{s (\cos(\theta^{(n)}_{{y_i}, i}) - m)}}{e^{s (\cos(\theta^{(n)}_{{y_i}, i}) - m)} + \sum_{j \neq y_i}{e^{s \cos(\theta^{(n)}_{j, i})}}}}},
\end{equation}
subject to
\begin{equation}
\begin{split}
  cos(\theta^{(n)}_j,i) &= {W_j}^Tx^{(n)}_i,
\end{split}
\end{equation}
where $N$ is the number of training images, $x^{(n)}_i$ is the $i$-th embedding vector extracted from night-style data corresponding to the ground truth class of $y_i$, the $W_j$ is the weight vector of the $j$-th class, and $\theta^{(n)}_j$ is the angle between $W_j$ and $x^{(n)}_i$. $s$ and $m$ are two hyper-parameters that control the weights of the intra-class distance and inter-class distance in the loss function, respectively.

Building on the previous notation, the softened probabilities obtained from the pre-trained model on the $i$-th daytime data can be expressed as:
\begin{equation}
\begin{split}
  p^{(d)}_i = \frac{e^{s (\cos(\theta^{(d)}_{{y_i}, i}) - m)}}{e^{s (\cos(\theta^{(d)}_{{y_i}, i}) - m)} + \sum_{j \neq y_i}{e^{s \cos(\theta^{(d)}_{j, i})}}}.
\end{split}
\end{equation}
Incorporating the above expression for softened probabilities $p^{(d)}_i$ and $p^{(n)}_i$ into the following equation:
\begin{equation}
\begin{split}
  L_{\mathrm{IKT}} = \frac{1}{N}\sum_{i}{D_{\mathrm{KL}}(p^{(d)}_i || p^{(n)}_i)},
\end{split}
\end{equation}
where $D_{\mathrm{KL}}$ represents the KL divergence. Finally, the final loss is:

\begin{equation}
\begin{split}
  L = L_{\mathrm{LMC}} + \alpha L_{\mathrm{IKT}},
\end{split}
\end{equation}
where $\alpha$ is a hyperparameter used to balance the two losses.

\subsection{Retrieval based on Partial Divide-and-Conquer}
\label{sec:pdc}

We have decomposed general VPR into two distinct sub-problems, Day-to-Day VPR and Night-to-Day VPR, addressing these challenges using two separate models. This approach is novel, and we explore its practicality and potential for optimization.

In real-world applications, computers can easily differentiate between day and night using straightforward methods that require only one-time operations. For instance, we can determine day and night simply by the system time of a device and approximate latitude and longitude information \footnote{\url{https://www.timeanddate.com/sun/italy/milan}}. Therefore, NocPlace involves only model parameter switching within the same architecture, without introducing additional inference overhead.

An inherent limitation of NocPlace is the requirement for computers to store two models and two databases. With the advent of an edge-cloud collaborative VPR system and the ease of scalable storage solutions in modern computing, this drawback becomes a manageable trade-off. However, if both the database and the model are on the edge, such as in decentralized autonomous robots, it would significantly consume storage resources. For instance, each image in the SF-XL test set is output as a feature vector of 512 dimensions, and the entire database size would be a staggering 5.7 GB. To overcome this challenge, we only use NocPlace to process the query image during the testing phase, while the database retains the features extracted by the pre-trained model. 
This section corresponds to the experimental setup in \cref{tab:ablation} regarding whether the original database is used.

\section{Experiments}
\label{sec:experiments}

In this section, we describe the implementation details, the test sets, and the evaluation metrics. Subsequently, we present a comprehensive range of quantitative experiments focusing on night-to-day visual place recognition, night-to-day visual localization (VL), and day-to-night image-to-image translation. Finally, extensive qualitative experimental results are presented in the supplementary material.

\subsection{Implementation details}

\noindent \textbf{The architecture of NocPlace.} The core idea of NocPlace revolves around transferring generated nocturnal knowledge and inherited day knowledge into existing models. Therefore, it is expected to be compatible with all previous methods and models, enhancing their performance in nocturnal scenarios. To illustrate the potential and scalability of NocPlace, we followed a methodology akin to CosPlace \cite{cosplace} and EigenPlaces \cite{eigenplaces}, leveraging a simple convolutional neural network architecture, specifically VGG-16 \cite{vgg16} and ResNet-50 \cite{resnet}. The architecture was connected to a GeM pooling layer \cite{dir} and a fully connected layer, with the dimensionality of the descriptors produced by NocPlace being determined by the specifications of the fully connected layer.

\noindent \textbf{The Training of NocPlace.} NocPlace employed CosPlace (512 dimensions) or EigenPlaces (2048 dimensions) as pre-trained models, requiring only one epoch of fine-tuning on SF-XL-NC with a learning rate of 1e-6. An important distinction is that NocPlace no longer utilizes any data augmentation techniques. The parameter $\alpha$ in the loss function can be adjusted by observing the initial loss of $L_{\mathrm{LMC}}$, but it does not require meticulous tuning. For instance, in training a model based on the ResNet-50 with an output dimension of 512, this parameter was set to 30.

\subsubsection{Night-to-Day Visual Localization.} We adopt the hierarchical localization framework proposed by \cite{hloc}, utilizing Superpoint \cite{superpoint} for local feature extraction and LightGLUE \cite{superglue, lightglue} for feature matching. The key modification we introduce lies in the VPR module, where we adjust the input image resolution to align it appropriately with the framework's predefined parameters.

\subsubsection{Day-to-Night Image-to-Image Translation.} We did not modify any training parameters and followed the official tutorial directly \cite{cut, negcut}. The difference in image resolution between NightCities and the VPR training sets is worth noting. To address this, we proportionally resize the NightCities images, keeping the shorter side at 640 pixels, and subsequently apply random cropping to achieve a size of 512$\times$512 during the training process. This approach ensures as much resolution consistency as possible between the I2I and VPR training sets. During the evaluation, we trained NEGCUT on multiple data sets mentioned in \cref{sec:experiments} and used NightStreets as the test set.

\subsection{Test sets}

\begin{table*}[t]
\begin{center}
\centering
\caption{\textbf{Overview of night VPR and VL test sets.} We can see that the test sets listed below includes various scales, scenarios, collection methods, and orientations.}
\label{tab:testsets}
\resizebox{\textwidth}{!}{
\begin{tabular}{l|ccccccc}
\toprule
\multirow{2}{*}{Dataset Name}  & SVOX & SF-XL & MSLS & Tokyo 24/7 & Tokyo 24/7 & Aachen v1.1 & RobotCar Seasons v2 \\
 & Night \cite{sxov} & Night \cite{reranking} & Night \cite{msls} & Night \cite{netvlad} & Sunset \cite{netvlad} & Night \cite{hloc} & Night \cite{robotcar}  \\
\hline
\# queries  & 823 & 466 & 55 & 105 & 105 & 192 & 438 \\
\# database  & 17k & 2.8M & 18.9k & 76k & 76k & 6.7k & 6.9k\\
Orientation& frontal-view &  multi-view &  frontal-view & multi-view & multi-view  & multi-view & frontal-view\\
Scenery & mostly urban & urban & urban & urban & urban & mostly urban & mostly urban \\
Acquisition Method & vehicles & handheld devices & vehicles & handheld devices & handheld devices & handheld devices & vehicles \\
\bottomrule
\end{tabular}}
\end{center}
\end{table*}

We conducted experiments on multiple nighttime VPR datasets, encompassing challenges frequently discussed in daytime VPR scenarios: large-scale, viewpoint variations, and scale changes. Some details are presented in \cref{tab:testsets}. Aside from reasonable modifications to Tokyo 24/7 \cite{netvlad} and MSLS \cite{msls}, the rest follow prior works \cite{3d_model, vpr_bench, vl_benchmarking, eigenplaces}.

\noindent \textbf{Tokyo 24/7} \cite{netvlad} is highly challenging in scale and has a three-year gap between query images and database images. Notably, the query images for each location are distributed across three time periods: daytime, sunset, and nighttime. To the best of our knowledge, few studies \cite{reranking, netvlad}  \textit{segment} this test set according to these time frames, somewhat overshadowing the impact of nighttime.

\noindent \textbf{MSLS} \cite{msls} is a crowdsourced dataset derived from dashcam recordings. We followed the validation set configuration from \cite{cosplace} and filtered out nighttime images. The experimental results for MSLS Night and the results on the public MSLS leaderboard \cite{msls} show significant differences, but the comparison is fair.

\subsection{Evaluation metrics} 

\noindent \textbf{Night-to-Day Visual Place Recognition.} For VPR datasets, our evaluation metric is recall@N using a 25-meter criterion. Specifically, this denotes the fraction of query instances where at least one among the top N predictions falls within a 25-meter proximity to the query's ground truth location, as commonly adopted in prior studies \cite{netvlad, 247_dataset, appsvr, cosplace, sare, sfrs}.

\begin{table}[!ht]
\centering
\caption{\textbf{Recall@1 / 5 / 10 on Night-to-Day VPR datasets.} The best results under the same model and the same data set are shown in red.}
\resizebox{\textwidth}{!}{
   \begin{tabular}{lccc|c|c|c|c|c}
    \toprule
    \multirow{2}{*}{Method}   &
    \multirow{2}{*}{Backbone} & 
    \begin{tabular}[c]{@{}l@{}}Feat.       \\ Dim.            \end{tabular} &
    \begin{tabular}[c]{@{}l@{}}Training    \\ Dataset         \end{tabular} &
    \begin{tabular}[c]{@{}l@{}}SVOX        \\ Night           \end{tabular} & 
    \begin{tabular}[c]{@{}l@{}}SF-XL       \\ Night           \end{tabular} & 
    \begin{tabular}[c]{@{}l@{}}MSLS        \\ Night           \end{tabular} &
    \begin{tabular}[c]{@{}l@{}}Tokyo 24/7  \\ Night           \end{tabular} &
    \begin{tabular}[c]{@{}l@{}}Tokyo 24/7  \\ Sunset          \end{tabular} \\
    \midrule
    \begin{tabular}[c]{@{}l@{}l@{}} NetVLAD \cite{netvlad} \\  SFRS \cite{sfrs} \\ CAE-VPR \cite{ye2023condition} \\CosPlace \cite{cosplace}   \\ EigenPlaces \cite{eigenplaces} \\ \textbf{NocPlace (Ours)}\end{tabular} & 
    \begin{tabular}[c]{@{}l@{}l@{}}   VGG-16 \\ VGG-16 \\ VGG-16 \\ VGG-16 \\ VGG-16 \\ VGG-16 \end{tabular} & 
    \begin{tabular}[c]{@{}l@{}l@{}}  4096 \\ 4096 \\ 4096 \\ 512 \\ 512 \\ 512 \end{tabular} & 
    \begin{tabular}[c]{@{}l@{}l@{}}  Pitts-30k \\ Pitts-30k \\ RobotCar \\ SF-XL \\ SF-XL \\ SF-XL-NC \end{tabular}& 
    \begin{tabular}[c]{@{}l@{}l@{}}  8.0 / 17.4 / 23.1  \\ 28.7 / 40.6 / 46.4 \\ 1.1 / 4.1 / 5.8 \\ 44.8 / 63.5 / 70.0 \\ 42.3 / 61.0 / 68.5 \\  \textcolor{red}{53.0} / \textcolor{red}{71.2} / \textcolor{red}{76.4} \end{tabular} & 
    \begin{tabular}[c]{@{}l@{}l@{}}  6.7 / 12.0 / 14.2  \\ 11.6 / 18.9 / 22.1 \\ 1.3 / 2.4 / 3.0 \\ 17.8 / 24.7 / 28.3 \\ 18.5 / 26.4 / 28.1 \\ \textcolor{red}{20.8} / \textcolor{red}{28.8} / \textcolor{red}{31.8} \end{tabular} &
    \begin{tabular}[c]{@{}l@{}l@{}}  1.8 / 1.8 / 3.6  \\ 1.8 / 7.3 / 14.5 \\ 1.8 / 1.8 / 1.8  \\ 10.9 / 16.4 / 29.1 \\ 1.8 / 12.7 / 18.2 \\ \textcolor{red}{29.1} / \textcolor{red}{50.9} / \textcolor{red}{56.4}\end{tabular} &
    \begin{tabular}[c]{@{}l@{}l@{}}  38.1 / 58.1 / 61.0  \\ 61.0 / 77.1 / 84.8 \\ 14.3 / 28.6 / 34.3 \\ 65.7 / 84.8 / 87.6  \\ 67.6 / 81.0 / 84.8 \\ \textcolor{red}{79.0} / \textcolor{red}{86.7} / \textcolor{red}{91.4}\end{tabular} &
    \begin{tabular}[c]{@{}l@{}l@{}}  81.9 / 89.5 / 90.5   \\ 87.6 / 91.4 / 91.4 \\ 21.9 / 32.4 / 38.1 \\ \textcolor{red}{89.5} / 90.5 / 93.3 \\ 87.6 / \textcolor{red}{94.3} / \textcolor{red}{97.1} \\ 88.6 / 91.4 / 94.3 \end{tabular}\\
    \midrule
    \begin{tabular}[c]{@{}l@{}l@{}} CosPlace \cite{cosplace} \\ Conv-AP \cite{gsv_cities}  \\ MixVPR \cite{ali2023mixvpr}  \\ EigenPlaces \cite{eigenplaces}  \\ \textbf{NocPlace (Ours)}     \end{tabular} &
    \begin{tabular}[c]{@{}l@{}l@{}} Res-50       \\ Res-50   \\ Res-50        \\ Res-50     \\  Res-50           \end{tabular} &
    \begin{tabular}[c]{@{}l@{}l@{}} 512          \\ 512   \\ 512            \\ 512        \\  512          \end{tabular} &
    \begin{tabular}[c]{@{}l@{}l@{}} SF-XL \\ GSV-Cities \\ GSV-Cities \\ SF-XL \\  SF-XL-NC      \end{tabular} & 
    \begin{tabular}[c]{@{}l@{}l@{}}  51.6 / 68.8 / 76.1 \\ 36.0 / 52.5 / 61.2 \\ 44.8 / 63.2 / 71.0 \\ 51.5 / 70.8 / 78.4 \\ \textcolor{red}{63.9} / \textcolor{red}{77.9} / \textcolor{red}{84.9}\end{tabular} &
    \begin{tabular}[c]{@{}l@{}l@{}} 23.8 / 29.0 / 31.5 \\ 4.5 / 7.9 / 10.1 \\ 10.7 / 18.2 / 20.0 \\ 22.3 / 29.8 / 32.8 \\ \textcolor{red}{26.6} / \textcolor{red}{32.0} / \textcolor{red}{34.5}  \end{tabular}& 
    \begin{tabular}[c]{@{}l@{}l@{}}  7.3 / 10.9 / 16.4 \\ 1.8 / 9.1 / 10.9 \\ 1.8 / 12.7 / 20.0 \\ 3.6 / 10.9 / 12.7 \\ \textcolor{red}{65.5} / \textcolor{red}{74.5} / \textcolor{red}{85.5}  \end{tabular} &
    \begin{tabular}[c]{@{}l@{}l@{}}  80.0 / 88.6 / 91.4 \\ 39.0 / 58.1 / 65.7 \\ 55.2 / 71.4 / 76.2 \\ 79.0 / 87.6 / 90.5 \\ \textcolor{red}{89.5} / \textcolor{red}{93.3} / \textcolor{red}{94.3} \end{tabular} &
    \begin{tabular}[c]{@{}l@{}l@{}}  91.4 / 97.1 / \textcolor{red}{99.0} \\ 67.6 / 80.0 / 88.6 \\ 82.9 / 93.3 / 95.2\\ \textcolor{red}{95.2} / \textcolor{red}{98.1} / 98.1 \\ 94.3 / 96.2 / 98.1  \end{tabular} \\
    \midrule
    \begin{tabular}[c]{@{}l@{}l@{}} CosPlace \cite{cosplace} \\ Conv-AP \cite{gsv_cities}  \\ EigenPlaces \cite{eigenplaces}  \\ \cite{eigenplaces} + ToDay GAN \cite{todaygan} \\ \textbf{NocPlace (Ours)}     \end{tabular} &
    \begin{tabular}[c]{@{}l@{}l@{}} Res-50  \\ Res-50  \\  Res-50   \\ Res-50 \\  Res-50           \end{tabular} &
    \begin{tabular}[c]{@{}l@{}l@{}} 2048  \\ 2048 \\ 2048    \\  2048 \\  2048          \end{tabular} &
    \begin{tabular}[c]{@{}l@{}l@{}} SF-XL \\ GSV-Cities \\ SF-XL \\ RobotCar \\  SF-XL-NC     \end{tabular} & 
    \begin{tabular}[c]{@{}l@{}l@{}}  50.8 / 67.6 / 74.8 \\ 37.9 / 57.1 / 65.4 \\ 58.8 / 76.9 / 82.6 \\ 13.0 / 27.9 / 36.0 \\ \textcolor{red}{75.6} / \textcolor{red}{87.5} / \textcolor{red}{91.4}  \end{tabular} &
    \begin{tabular}[c]{@{}l@{}l@{}}   23.6 / 29.2 / 32.8  \\ 5.6 / 10.1 / 12.0 \\ 23.6 / 30.9 / 34.5 \\ 3.9 / 8.6 / 10.9 \\ \textcolor{red}{28.3} / \textcolor{red}{35.8} / \textcolor{red}{39.7}\end{tabular} &
    \begin{tabular}[c]{@{}l@{}l@{}}  5.5 / 10.9 / 16.4 \\ 1.8 / 9.1 / 18.2 \\ 3.6 / 10.9 / 10.9 \\ 12.7 / 20.0 / 25.5 \\ \textcolor{red}{69.1} / \textcolor{red}{76.4} / \textcolor{red}{80.0}  &
    \end{tabular} &
    \begin{tabular}[c]{@{}l@{}l@{}}  78.1 / 89.5 / 91.4  \\ 47.6 / 60.0 / 66.7 \\ 83.8 / 92.4 / 94.3 \\ 22.9 / 40.1 / 48.7\\  \textcolor{red}{91.4} / \textcolor{red}{94.3} / \textcolor{red}{96.2 } \end{tabular} & 
    \begin{tabular}[c]{@{}l@{}l@{}}  89.5 / 94.3 / 96.2 \\ 77.1 / 86.7 / 89.5 \\ \textcolor{red}{96.2} / 98.1 / 98.1 \\ 8.6 / 17.2 / 24.8 \\  93.3 / \textcolor{red}{99.0} / \textcolor{red}{99.0}  \end{tabular} \\
    \bottomrule
    \end{tabular}}
\label{tab:vpr_comparsion}
\end{table}

\noindent \textbf{Night-to-Day Visual Localization.}  We evaluate VPR through the localization success rate under different recall@N metrics \cite{hloc, lightglue, vl_benchmarking, long_vl}. It is crucial to note that the VL metrics demand more rigorous precision than VPR metrics.

\noindent \textbf{Day-to-Night Image-to-Image Translation.} Our evaluation of translated images encompasses two primary facets: \textit{realism} and \textit{faithfulness} \cite{unpaired, cut, negcut}. Realism quantitatively assesses the disparity between the target and translated image distributions. Faithfulness gauges the extent of congruence between an original image and its translation. We leverage the established Frechet Inception Score (FID) for a precise measure of realism. In measuring Faithfulness, we utilize metrics like the pixel-wise $L_{2}$ distance, peak-signal-to-noise ratio (PSNR), and the structural similarity index measure (SSIM) for each paired input-output.

\subsection{Comparison with previous work}

We have conducted extensive comparisons of numerous methods on various datasets, an endeavor unprecedented in nighttime VPR, as shown in \cref{tab:vpr_comparsion} and \cref{tab:vl_comparsion}. These methods range from the classical NetVLAD\cite{netvlad}, To-Day GAN \cite{todaygan} to more recent contributions over the past three years, including SFRS \cite{sfrs}, CosPlace \cite{cosplace}, MixVPR \cite{ali2023mixvpr}, CAE-VPR \cite{ye2023condition}, Conv-AP \cite{gsv}, and EigenPlaces \cite{eigenplaces}. We replicated their open-source models and code for our experiments. It is worth noting that these methods were also tested on SVOX Night in \cite{eigenplaces}, and our results are consistent with theirs. The experimental results can be distilled into the following key observations:

\begin{itemize}

\item NocPlace is \textit{the undisputed winner} in nighttime scenarios. Its performance enhancements are substantial, even compared to EigenPlaces (ICCV'2023).

\item SF-XL Night (CVPRW'2023) is the most challenging nighttime VPR test set. On this test set, the Recall@1 of NocPlace is \textit{4.7}\% higher than the Recall@1 of EigenPlaces.

\item On both SVOX Night and Tokyo 24/7 Night, NocPlace even shows performance improvements \textit{across} architectures and feature dimensions. For example, VGG16-based NocPlace excels over ResNet50-based EigenPlaces.

\item The advantages of NocPlace are magnified on MSLS Night due to the limited number of query images (less than 100) and redundancy (sequential frames).

\item  The sunset scenes in Tokyo 24/7 present a mix of numerous daytime and fewer nighttime images, which exposes the degradation of NocPlace during daylight conditions. As long as model switching can be considered as described in \cref{sec:pdc}, this degradation will not occur in \textit{practice}.

\item In the case of high-precision positioning and small-scale databases, the challenge and importance of VPR are weakened. On Aachen v1.1, the advantage of NocPlace became less clear, trailing slightly behind SFRS (ECCV'2020) by 1\%.

\item On RobotCar Seasons v2 (RS-v2), ToDay GAN enhanced the performance of EigenPlace, and CAE-VPR achieved the best performance. However, they show poor generalization on other test sets, which supports the viewpoints made in \cref{sec:introduction} and \cref{sec:related_work}. In particular, ToDay GAN needs to execute two models in series when performing a VPR task, which consumes additional computing resources. RS-v2 and the training set of CAE-VPR are the same part of the original RobotCar. So, the results of CAE-VPR may be unfair to NocPlace and other methods on RS-v2.

\end{itemize}

\begin{table*}[t]
\centering
\caption{\textbf{Localization success rate at different localization thresholds, recall@N, and datasets.} \cite{long_vl}.  The localization thresholds are (0.25m, 2\degree) / (0.5m, 5\degree) / (5m, 10\degree). Specifically, recall numbers are set to 1, 5, 10. The best results for the same data set and the same threshold are shown in red.}
\resizebox{\textwidth}{!}{
   \begin{tabular}{lcc|c|c|c|c|c|c}
    \toprule
    \multirow{2}{*}{Method} &
    \multirow{2}{*}{Backbone} & 
    \multirow{2}{*}{\begin{tabular}[c]{@{}l@{}}Feat. \\ Dim.\end{tabular}} & 
    \multicolumn{3}{c|}{Aachen v1.1 Night} & \multicolumn{3}{c}{RobotCar Seasons v2 Night} \\
    & & &  Recall@1 & Recall@5 & Recall@10  &  Recall@1 & Recall@5 & Recall@10 
    \\
    \midrule
    \begin{tabular}[c]{@{}l@{}l@{}}NetVLAD\cite{netvlad} \\ SFRS\cite{sfrs}\\ CAE-VPR \cite{ye2023condition}\\ Conv-AP \cite{gsv_cities} \\ CosPlace \cite{cosplace}\\ EigenPlaces \cite{eigenplaces} \\ \cite{eigenplaces} + ToDay GAN \cite{todaygan} \\ 
    \textbf{NocPlace (Ours)} \end{tabular} &
    \begin{tabular}[c]{@{}l@{}l@{}} VGG16       \\ VGG16 \\ VGG16 \\ ResNet-50  \\ ResNet-50        \\ ResNet-50  \\ ResNet-50  \\  ResNet-50           \end{tabular} &
    \begin{tabular}[c]{@{}l@{}l@{}} 4096  \\ 4096   \\ 4096 \\  2048    \\   2048  \\  2048 \\ 2048 \\ 2048          \end{tabular} &
    
    \begin{tabular}[c]{@{}l@{}l@{}} 63.4 / 80.6 / 88.0 \\  \textcolor{red}{69.6} / 85.3 / 95.3 \\ 	46.6 / 55.5 / 64.4 \\ 57.6 / 77.0 / 85.3 \\ 66.5 / 84.3 / \textcolor{red}{95.8} \\ 67.5 / 83.2 / 93.7 \\ 22.5 / 31.4 / 38.2 \\ 68.6 / \textcolor{red}{85.9} / \textcolor{red}{95.8}\end{tabular} & 
    
    \begin{tabular}[c]{@{}l@{}l@{}} 68.6 / 87.4 / 97.4  \\ 71.2 / 87.4 / 97.4 \\ 57.6 / 70.2 / 78.5 \\ 71.2 / 85.9 / 94.8 \\ 71.2 / 86.9 / \textcolor{red}{97.9} \\ \textcolor{red}{72.8} / 88.5 / \textcolor{red}{97.9} \\ 37.7 / 50.3 / 57.6 \\ \textcolor{red}{72.8} / \textcolor{red}{89.0} / 96.9  \end{tabular} &
    
    \begin{tabular}[c]{@{}l@{}l@{}} 70.7 / 85.9 / 95.3 \\ 71.7 / 88.0 / 98.4 \\ 60.7 / 77.0 / 83.8 \\ \textcolor{red}{74.9} / 89.0 / 96.9 \\ 73.8 / 88.0 / 97.9 \\ 73.8 / 88.0 / 97.9 \\ 42.9 / 56.0 / 64.9 \\ \textcolor{red}{74.9} / \textcolor{red}{90.1} / \textcolor{red}{99.0} \end{tabular} &
    
    \begin{tabular}[c]{@{}l@{}l@{}} 11.5 / 22.1 / 30.1 \\ 18.1 / 35.8 / 48.2 \\ \textcolor{red}{25.7} / \textcolor{red}{55.8} / \textcolor{red}{67.7} \\ 21.2 / 38.5 / 51.3 \\ 14.6 / 26.1 / 40.7 \\ 14.2 / 23.9 / 38.5 \\ 14.6 / 25.2 / 40.7 \\ 22.1 / 41.6 / 58.8 \end{tabular} & 

    \begin{tabular}[c]{@{}l@{}l@{}} 18.1 / 38.9 / 50.9 \\ 21.7 / 50.0 / 63.7 \\ \textcolor{red}{34.1} / \textcolor{red}{69.5} / \textcolor{red}{81.0} \\ 25.7 / 56.2 / 67.3 \\ 19.0 / 34.1 / 48.2 \\16.8 / 31.0 / 46.0 \\ 19.9 / 39.4 / 54.0 \\ 23.9 / 54.4 / 72.6 \end{tabular} &
    
    \begin{tabular}[c]{@{}l@{}l@{}} 21.2 / 46.9 / 58.0 \\ 23.5 / 59.3 / 70.8 \\ \textcolor{red}{32.7} / \textcolor{red}{72.6} / \textcolor{red}{84.1} \\ 29.2 / 62.4 / 75.7 \\ 19.0 / 36.3 / 51.3 \\ 18.1 / 36.7 / 50.4 \\ 27.4 / 48.2 / 61.5 \\ 28.8 / 58.8 / 79.2  \end{tabular} \\

    \bottomrule
  \end{tabular}
}
\label{tab:vl_comparsion}
\end{table*}

\subsection{Ablations}

\begin{table}[t]
\caption{\textbf{Ablations.} This table shows the Recall@1/5/10 obtained with a ResNet-50 with output dimensionality 512 on two datasets. IKT means Inherited Knowledge Transfer, and OD means using the original database.}
\centering
\begin{tabular}{cccc|c|c}
\toprule
\multicolumn{2}{c}{Training set} & \multirow{2}{*}{IKT} & \multirow{2}{*}{OD}  & \multicolumn{1}{c}{Tokyo 24/7}   & \multicolumn{1}{c}{SVOX}\\
Day    & Night    &                      &                               & \multicolumn{1}{c}{Night} & \multicolumn{1}{c}{Night}\\
\hline
\checkmark &            &            & \checkmark  & 80.0 / 88.6 / 92.4  & 51.6 / 68.8 / 76.1 \\
\midrule
\checkmark & \checkmark &            & \checkmark  & 83.8 / 94.3 / 94.3  & 62.3 / 77.4 / 81.9 \\
           & \checkmark &            & \checkmark  & 83.8 / 93.3 / 96.2  & 63.2 / 78.7 / 84.8 \\
           & \checkmark & \checkmark & \checkmark  & 89.5 / 93.3 / 94.3  & 63.9 / 77.9 / 84.9 \\
\midrule
\checkmark & \checkmark &            &             & 86.7 / 92.4 / 92.4  & 65.6 / 80.8 / 85.3 \\

           & \checkmark &            &             & 87.6 / 93.3 / 92.4  & 66.6 / 80.9 / 87.1 \\

           & \checkmark & \checkmark &             & 91.4 / 93.3 / 93.3  & 67.3 / 81.0 / 86.6 \\
\bottomrule
\end{tabular}
\label{tab:ablation}
\end{table}

We conducted ablation experiments on the ResNet-50 architecture with an output dimension of 512 for NocPlace, as shown in \cref{tab:ablation}. It is evident that generating nighttime data and incorporating the IKT loss results in significant performance improvements. Here, OD means using the original database, which also means the partial divide-and-conquer-based retrieval mentioned before. By using OD, we sacrificed accuracy to maintain the practicality of NocPlace. This implies that without using OD, NocPlace in \cref{tab:vpr_comparsion} and \cref{tab:vl_comparsion} could achieve even better experimental results.

\subsection{Comparision with previous datasets}

\begin{table}[t]
\caption{\textbf{Quantitative comparison of NEG-CUT \cite{negcut} on different night-to-day datasets.} The best results are shown in red. Only the day2night dataset uses the pre-trained model of Pix2Pix \cite{pix2pix}.}
\centering
\begin{tabular}{llcccc}
\toprule
Dataset & Method & FID $\downarrow$ & L2 $\downarrow$ & PSNR $\uparrow$ & SSIM $\uparrow$\\
\midrule
day2night \cite{day2night} & Pix2Pix & 243.90 & 164.79 & 8.74 & 0.21 \\
LOL \cite{lol_dataset} & NEG-CUT & 99.62 &  191.99  &  7.40 &  0.14 \\
RobotCar \cite{robotcar} & NEG-CUT & 148.88 & 156.29 & 9.29 & 0.32  \\
NightStreets & NEG-CUT  & 63.54 & 121.62 & 11.75 & 0.60  \\
NightCities (Ours) & NEG-CUT  & \textcolor{red}{62.21} & \textcolor{red}{98.12} & \textcolor{red}{13.98} & \textcolor{red}{0.75} \\
\bottomrule
\end{tabular}
\label{tab:i2i_datasets}
\end{table}

We focus on the quality of day-to-night datasets and the performance of image-to-image methods. We discuss the former here and the latter in the supplementary material. To objectively assess these datasets, we trained NEG-CUT on the previously mentioned and proposed datasets. Only the day2night dataset \cite{day2night} uses the pre-trained model of Pix2Pix \cite{pix2pix} because it is a paired day-night dataset. The test set uses NightStreets to comply with the principle of single-variable experiments. As shown in the \cref{tab:i2i_datasets}, the model trained on NightCities achieved the best evaluation metrics commonly used in the I2I translation domain, even surpassing the model trained on NightStreets. Considering that NightStreets is reorganized from the VPR test set, these results also show that the model trained on NightCities can generate data that is more similar to the test data.

\section{Conclusion}
\label{sec:conclusion}

In this paper, we introduce a novel training and testing strategy designed for VPR to tackle the challenges of cross-domain retrieval and localization between day and night scenarios. With the proposed unpaired day-night urban scene dataset, we transfer the appearance knowledge of nighttime to an existing large-scale daytime VPR dataset. By minimizing the classification loss in the target domain and the domain discrepancy between source and target domains, we further enhance the performance of night-to-day VPR. Considering practical use, a retrieval strategy based on partial divide-and-conquer was proposed for the first time. Extensive experiments conducted on comprehensive nighttime VPR test sets substantiate our contributions. As a straightforward and scalable approach, NocPlace addresses the degradation observed in existing VPRs during night-to-day recognition and holds promise for improving future VPRs and other downstream tasks in computer vision.


\section{Overview}
\label{sec:overview}

The table of contents of the supplementary content is as follows.

\begin{itemize}
    
    \item Discussions
    \begin{itemize}
        \item The importance of Night-to-Day VPR
        \item The novelty of generative knowledge transfer
        \item The choice of backbones and methods
        \item The choice of test sets
    \end{itemize}

    \item Experiments
    \begin{itemize}
        \item Comparisons with SoTA Methods (2024.03)
        \item Comparing I2I translation methods
        \item Qualitative results
    \end{itemize}
    
    \item Datasets
    \begin{itemize}
        \item The proposed datasets
        \item The Datasets of VPR
    \end{itemize}

\end{itemize}

\section{Discussions}
\label{sec:discussions}

\subsection{The importance of Night-to-Day VPR.} Night-to-day VPR, as a sub-problem of VPR, may make readers think it is unimportant. We provide the following insights:

\begin{itemize}
    \item Many studies \cite{dannet, lengyel2021zero, risp, todaygan, gan_ss} have shown that nighttime problems are equally important as daytime problems. For practical, we do not want intelligent robots and AR glasses to be available only during the day.
    \item While LiDAR is a nighttime alternative to cameras, it costs much more than the latter, is much larger, and does not provide color information.
    \item The nighttime problem differs other fundamental problems (e.g., visual scaling, occlusion) in that it can be predicted when it will occur and decomposed from the parent problem. However, researchers always try to solve it in independent systems. The divide-and-conquer strategy used here is novel.
\end{itemize}

\subsection{The novelty of generative knowledge transfer.} The simplicity of generative knowledge transfer (GKT) may lead some readers to question its novelty. We provide the following explanations:

\begin{itemize}

    \item ToDay GAN \cite{todaygan} is proposed to generate daytime images for the retrieval phase of VPR, while NocPlace reverses the idea of ToDay GAN and generates night images for the training phase of VPR. If a study like NocPlace existed, it would inevitably cite To-Day GAN. While researching NocPlace, we searched Google Scholar many times and found that no one had implemented this simple and seemingly uninnovative idea.

    \item DSA (ICCV'2023) \cite{dsa} is a recent contemporaneous work with NocPlace, which also utilizes an image generation network to obtain night-style training images. Experiments about DSA are supplemented in \cref{tab:vpr_comparsion_supp}.

    \item The fact that some experienced researchers \footnote{\url{https://github.com/AAnoosheh/ToDayGAN/issues/10}} consider the idea of GKT to be infeasible suggests that it has not been successfully implemented or proven effective before.

    \item Generative knowledge transfer is straightforward, but its implementation relies on high-quality images and advanced image generation methods. We solved the first problem and made many attempts at the latter.
    
\end{itemize}

\subsection{The choice of backbones and methods.}
\label{backbones}
Although the reasons for choosing backbones and methods have been stated in the paper, readers may still have concerns. We provide the following additional explanations:

\begin{itemize}
    \item The scaling law (OpenAI, 2020) \cite{kaplan2020scaling} is a basic law of deep learning. As shown in previous experiments, improving the model architecture or increasing the descriptor dimension can correspondingly improve the results, demonstrating the scalability of NocPlace. Experienced researchers can foresee the impact of foundation models. Moreover, foundation models for VPR can be studied as an independent research problem (e.g., AnyLoc, CricaVPR, and SALAD).
    \item We highly recognize the performance of AnyLoc, CricalVPR, and SALAD in public data sets and the inspiration they bring to the community. However, VPR is one of the technologies used in robots and AR glasses, which have stringent requirements for power, computing, and latency caused by limited resources (NeuroGPR \cite{yu2023brain}, Science Robotics'2023).
    \item NocPlace belongs to the one-stage VPR method, so there is no comparison experiment with the two-stage VPR method. In fact, NocPlace outperforms some two-stage state-of-the-art methods.
\end{itemize}

\subsection{The choice of test sets.} Although the experiments in this article cover many test sets, two test sets are still missing.  We provide the following explanations:

\begin{itemize}
    \item \textbf{Gardens Point \cite{gardenspoint} (2014) .} The official link is no longer available to obtain the Gardens Point dataset, and there is a community version \footnote{\url{https://github.com/aghagol/loop-detection/issues/2}}. However, this dataset has become unchallenging due to its size (200 daytime and 200 nighttime images).

    \item \textbf{Alderley \cite{seqslam} (2012).} The official link is no longer available to obtain the Alderley dataset \footnote{\url{https://open.qcr.ai/dataset/alderley/}}, and we did not find any community versions.
\end{itemize}

\section{Experiments}
\label{sec:exp}

\subsection{Comparing with SoTA Methods (2024)}
\label{sota}

We give the following observations:

\begin{itemize}

    \item On Tokyo 24/7 and MSLS Night, the Recall@1 of NocPlace is higher than the Recall@1 of CricaVPR (CVPR'2024) and AnyLoc (RAL'2024).
    \item NocPlace is far better than DSA (ICCV'2023) and EigenPlaces (ICCV'2023) on all data sets.
    \item SALAD (CVPR'2024) achieves the best results on the three data sets.
    \item CricaVPR, SALAD, and AnyLoc all use ViT-based DINO-v2 \cite{oquab2023dinov2} as feature extractors, and their feature dimensions are much higher than NocPlace.
    \item The parameters shown in \cref{tab:vpr_comparsion_supp} are the parameters of the backbone. NocPlace is the model with the smallest parameters and the fastest feature extraction speed.
\end{itemize}

\begin{table}[t]
\centering
\caption{\textbf{Recall@1 / 5 / 10 on Night-to-Day VPR datasets.}}
\resizebox{\textwidth}{!}{
   \begin{tabular}{lcc|c|c|c|c|c|c}
    \toprule
    \multirow{2}{*}{Method}   &
    \multirow{2}{*}{Backbone} & 
    \begin{tabular}[c]{@{}l@{}}Feat.       \\ Dim.            \end{tabular} &
    \begin{tabular}[c]{c}Params $\downarrow$  \\  (M)               \end{tabular} &
    \begin{tabular}[c]{c}Speed  $\uparrow$ \\   (it/s)              \end{tabular}  &
    \begin{tabular}[c]{@{}l@{}}SVOX        \\ Night           \end{tabular} & 
    \begin{tabular}[c]{@{}l@{}}Tokyo 24/7        \\            \end{tabular} & 
    \begin{tabular}[c]{@{}l@{}}MSLS        \\ Night           \end{tabular} &
    \begin{tabular}[c]{@{}l@{}} SF-XL\\  Night               \end{tabular} \\
    \midrule
    \begin{tabular}[c]{@{}l@{}l@{}}  DSA \cite{dsa} \\ DSA \cite{dsa} \\EigenPlaces  \cite{eigenplaces}  \\ NocPlace     \end{tabular} &
    \begin{tabular}[c]{@{}l@{}l@{}} VGG-16 \\ Res-101 \\  Res-50  \\  Res-50           \end{tabular} &
    \begin{tabular}[c]{@{}l@{}l@{}}  512 \\ 2048 \\ 2048  \\  2048  \end{tabular} &
    \begin{tabular}[c]{c}  138 \\ 45 \\ 24 \\  24  \end{tabular} &
    \begin{tabular}[c]{c}  58.86\\63.30 \\ 198.90 \\ 198.90   \end{tabular} &
    \begin{tabular}[c]{@{}l@{}l@{}}  8.3 / 19.0 / 23.1 \\10.4 / 20.0 / 25.3\\  58.8 / 76.9 / 82.6 \\ 75.6 / 87.5 / 91.4 \end{tabular} &
    \begin{tabular}[c]{@{}l@{}l@{}}  61.3 / 74.9 / 80.6 \\ 69.5 / 83.5 / 87.3\\ 93.0 / 96.2 / 97.5 \\  95.2 / 96.8 / 98.1  \end{tabular} &
    \begin{tabular}[c]{@{}l@{}l@{}}  3.6 / 3.6 / 7.3  \\ 3.6 / 5.5 / 9.1 \\ 3.6 / 10.9 / 10.9 \\  69.1 / 76.4 / 80.0  &
    \end{tabular} &
    \begin{tabular}[c]{@{}l@{}l@{}}   ----- / ----- / -----\\ ----- / ----- / ----- \\ 23.6 / 30.9 / 34.5 \\ 28.3 / 35.8 / 39.7 \end{tabular} \\
    \midrule
    \begin{tabular}[c]{@{}l@{}l@{}} AnyLoc \cite{keetha2023anyloc} \\ SALAD \cite{Izquierdo_CVPR_2024_SALAD} \\ CricaVPR \cite{lu_crica}    \end{tabular} &
    \begin{tabular}[c]{@{}l@{}l@{}} DINOv2-g \\ DINOv2-B \\ DINOv2-B      \end{tabular} &
    \begin{tabular}[c]{@{}l@{}l@{}} 49152  \\ 8448 \\ 4096   \end{tabular} &
    \begin{tabular}[c]{c}  1100 \\ 86 \\86   \end{tabular} &
    \begin{tabular}[c]{c}  3.24 \\ 30.40 \\56.24   \end{tabular} &
    \begin{tabular}[c]{@{}l@{}l@{}}  77.5 / 90.6 / 94.5 \\ 97.0 / 99.0 / 99.1 \\ 85.1 / 95.0 / 96.7 \end{tabular} &
    \begin{tabular}[c]{@{}l@{}l@{}}  91.7 / 97.5 / 98.7  \\97.5 / 98.4 / 98.4 \\ 93.0 / 97.5 / 98.1 \end{tabular} &
    \begin{tabular}[c]{@{}l@{}l@{}}  3.6 / 5.5 / 16.4 \\ 90.9 / 100.0 / 100.0 \\50.9 / 78.2 / 81.8 \end{tabular} &
    \begin{tabular}[c]{@{}l@{}l@{}}   ----- / ----- / ----- \\ ----- / ----- / ----- \\ ----- / ----- / -----  \end{tabular} \\
    \midrule
    \begin{tabular}[c]{@{}l@{}l@{}} SALAD-Noc    \end{tabular} &
    \begin{tabular}[c]{@{}l@{}l@{}} DINOv2-B     \end{tabular} &
    \begin{tabular}[c]{@{}l@{}l@{}} 8448         \end{tabular} &
    \begin{tabular}[c]{c}           86           \end{tabular} &
    \begin{tabular}[c]{c}           30.40        \end{tabular} &
    \begin{tabular}[c]{@{}l@{}l@{}} \textcolor{blue}{96.4} / 99.0 / 99.1 \end{tabular} &
    \begin{tabular}[c]{@{}l@{}l@{}} 97.5 / 98.4 / \textcolor{red}{98.7} \end{tabular} &
    \begin{tabular}[c]{@{}l@{}l@{}} \textcolor{red}{94.9} / \textcolor{blue}{98.2} / 100.0 \end{tabular} &
    \begin{tabular}[c]{@{}l@{}l@{}}   ----- / ----- / -----  \end{tabular} \\
    \bottomrule
    \end{tabular}}
\label{tab:vpr_comparsion_supp}
\end{table}

\subsection{Comparing I2I translation methods}

\begin{table}[t]
\caption{Comparsion of different unpaired I2I methods}
\centering
\begin{tabular}{llllll}
\toprule
Method & FID $\downarrow$ & L2 $\downarrow$ & PSNR $\uparrow$ & SSIM $\uparrow$& Avg Time $\downarrow$ \\
\midrule
CycleGAN \cite{unpaired} & 63.93 & \textbf{117.59} & \textbf{12.88} & \textbf{0.68} &  10 ms  \\
CUT \cite{cut}   & \textbf{63.20} & 129.47 & 11.12 & 0.60 & \textbf{3.2 ms} \\
NEGCUT \cite{negcut}  & 63.54 & 121.62 & 11.75 & 0.60 & \textbf{3.2 ms}   \\
ESGDE \cite{egsde} & - & - & - & - &  62000 ms  \\
\bottomrule
\end{tabular}
\label{tab:i2i_methods}
\end{table}

We trained a range of state-of-the-art unpaired I2I translation methods on NightStreets, as detailed in Tab. \ref{tab:i2i_methods}. Interestingly, the disparities in outcomes among these methods are considerably less pronounced than those observed on other public datasets. However, one aspect is clear: the increasingly popular Diffusion Model, though capable of generating high-quality images, is notably time-intensive. For example, processing a training set of 100,000 VPR images with ESGDE \cite{egsde} on a 3090 Ti GPU requires approximately 72 days. Given that our primary research focuses on VPR, we did not extensively dwell on these issues and adopted NEG-CUT \cite{negcut} for our study.

\subsection{Qualitative results}

\begin{figure*}[t] 
\begin{center}
{\includegraphics[width=\textwidth] {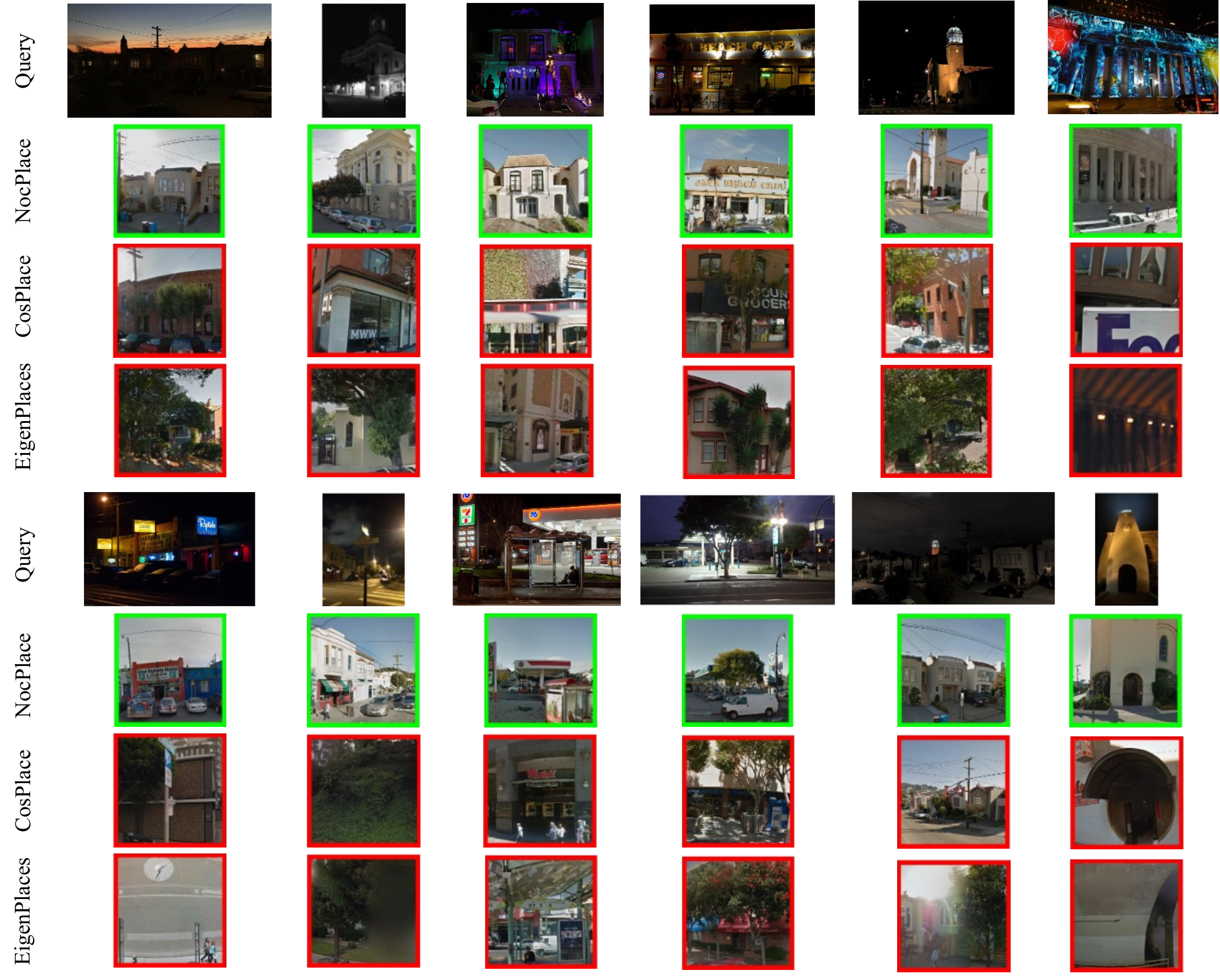}}
\end{center}
   \caption{\textbf{Qualitative results on the SF-XL Night dataset.} Each column represents a query (first and fourth rows) and the first predicted result from the database. We can see that NocPlace handles extreme lighting variations better than previous methods.}
\label{fig:demo_sf}
\end{figure*}

\begin{figure*}[t] 
\begin{center}
{\includegraphics[width=\textwidth] {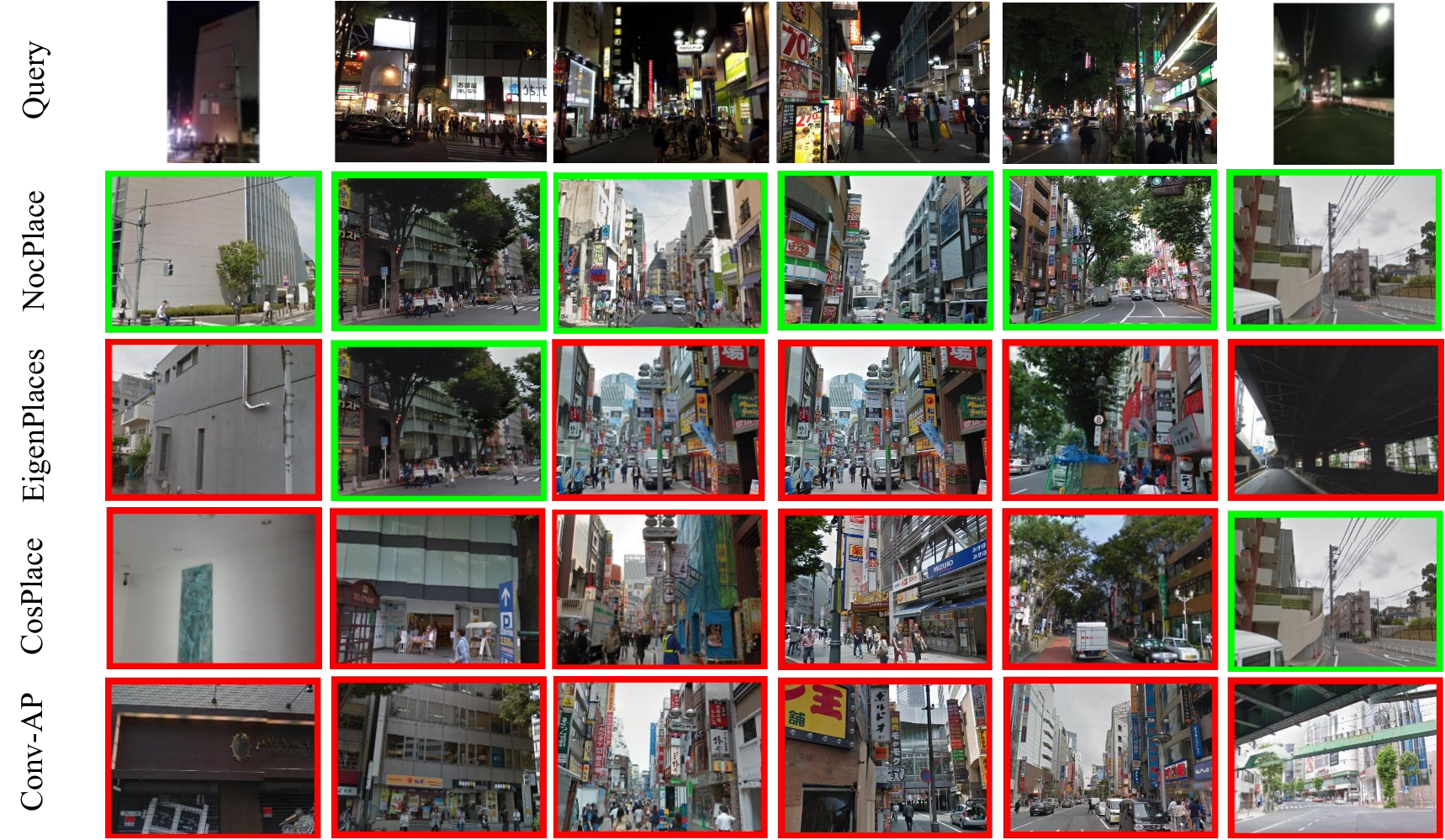}}
\end{center}
   \caption{\textbf{Qualitative results on the Tokyo 24/7 dataset.} Each column represents a query (first row) and the first predicted result from the database. We can see that NocPlace handles extreme lighting variations better than previous methods.}
\label{fig:demo_tokyo}
\end{figure*}

Some qualitative results are shown in Fig \ref{fig:demo_sf} and Fig. \ref{fig:demo_tokyo}. The figure allows us to understand the strengths of NocPlace more visually and intuitively. 

\section{Datasets}
\label{sec:datasets}

\subsection{The proposed datasets}

\subsubsection{NightCities}

\begin{figure}[t] 
\begin{center}
{\includegraphics[width=\textwidth] {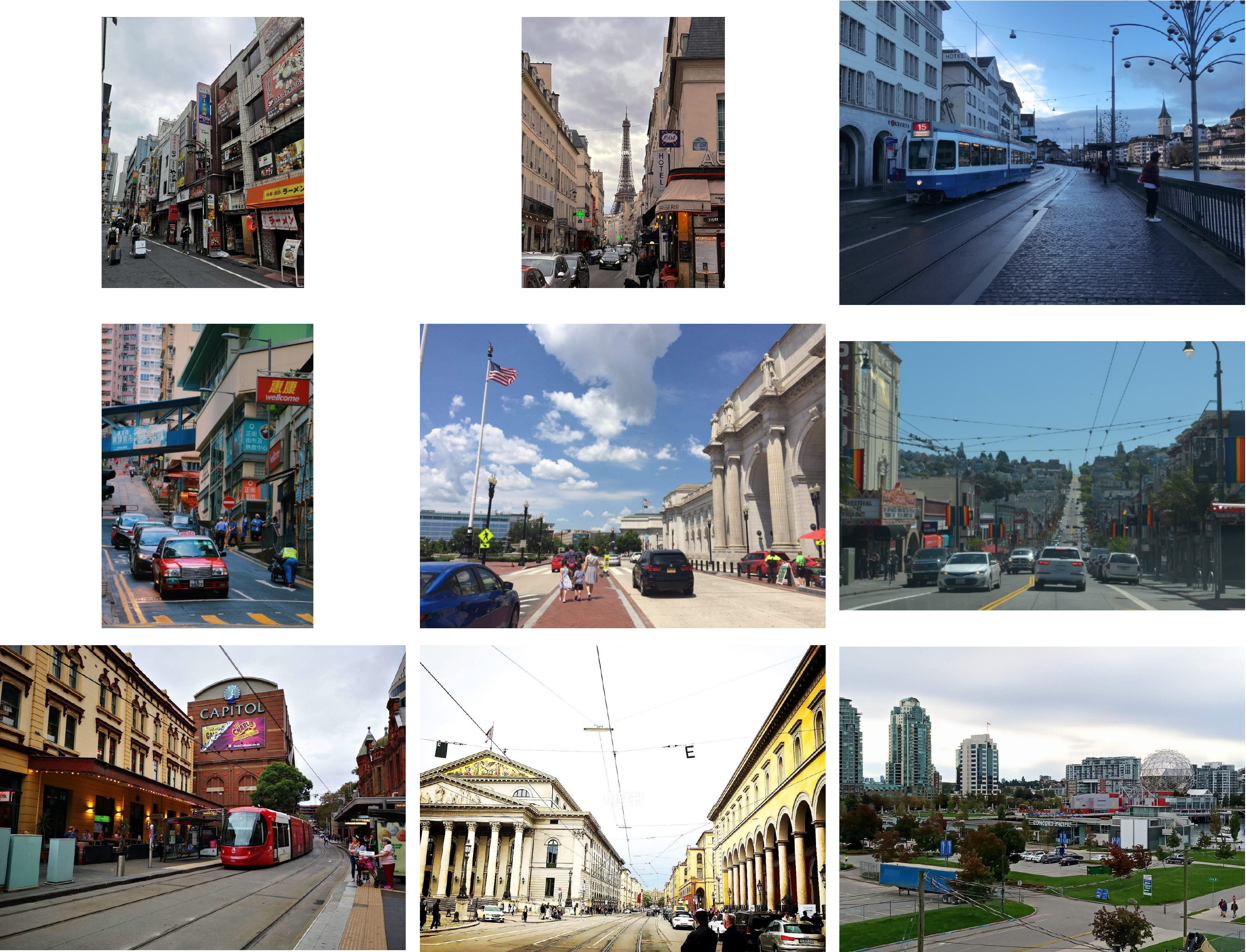}}
\end{center}
   \caption{\textbf{Daytime examples of the NightCities dataset.} We supplement the daytime images of the NightCities dataset here. From left to right, top to bottom, they represent Tokyo, Paris, Zurich, Hong Kong, Washington, San
   Francisco, Sydney, Munich, and Vancouver.}
\label{fig:nightcities_daytime}
\end{figure}

\begin{figure*}[t] 
\begin{center}
{\includegraphics[width=\textwidth] {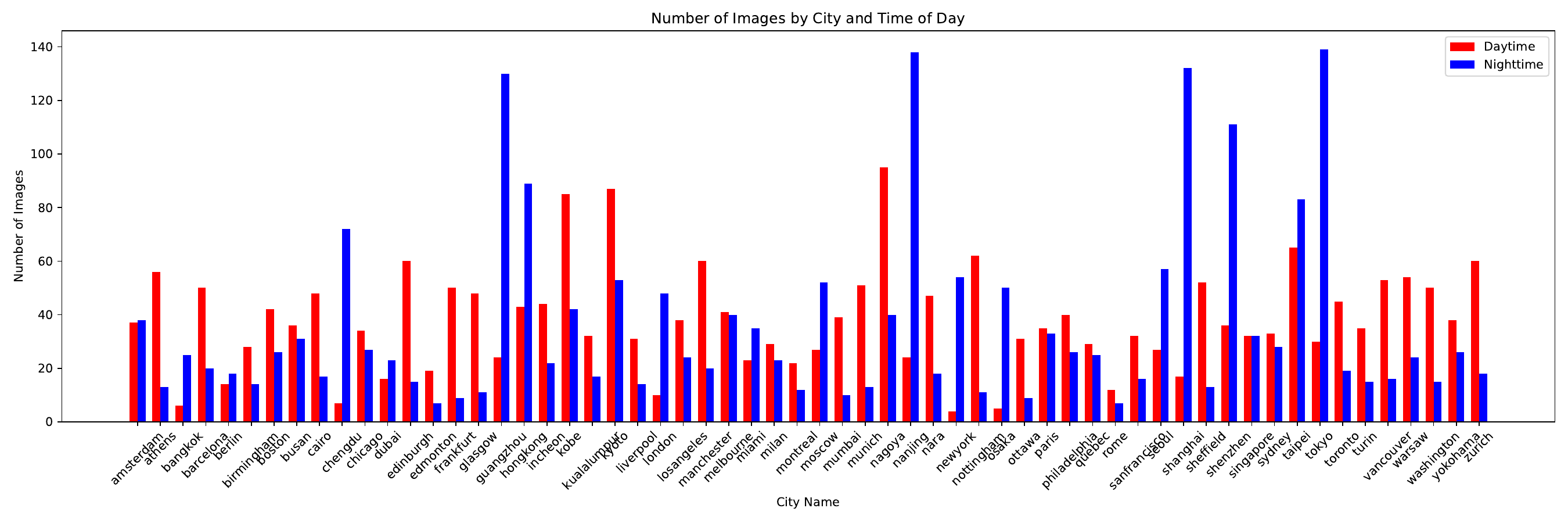}}
\end{center}
   \caption{\textbf{Histogram of the number of photos for each city.} We have collected daytime and nighttime street view images of 60 cities and verified their accuracy manually. Over the next 1-3 years, we will continue to maintain and expand this dataset.}
\label{fig:hist}
\end{figure*}

We supplement the NightCities with daytime images and category directories, as shown in Fig. \ref{fig:nightcities_daytime} and Fig. \ref{fig:hist}.

\subsubsection{SF-XL-NC}

\begin{figure*}[t] 
\begin{center}
{\includegraphics[width=\textwidth] {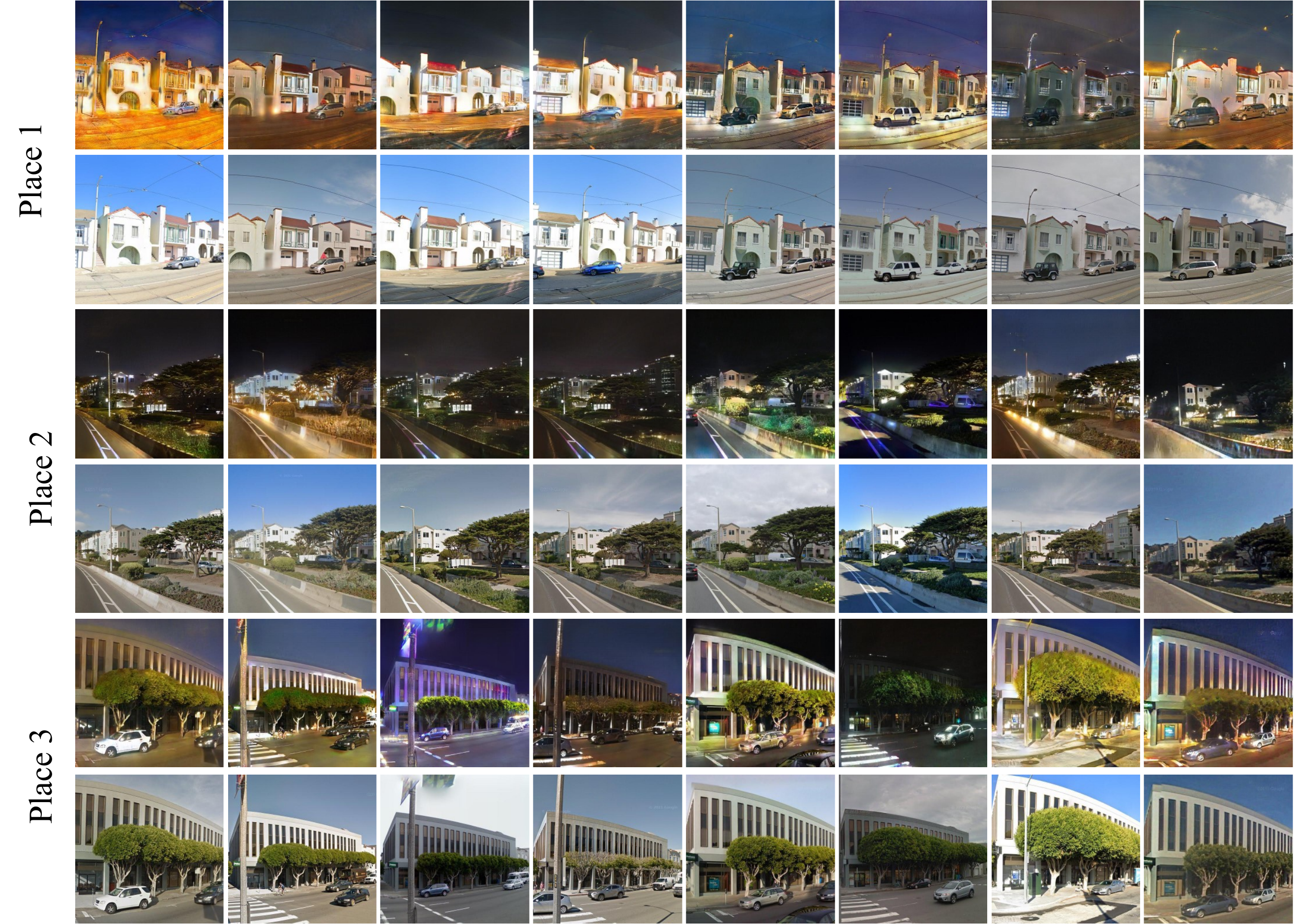}}
\end{center}
   \caption{\textbf{Partial instances of the SF-XL-NC dataset.} The second, fourth, and sixth rows are images from three locations in SF-XL, while the first, third, and fifth rows are the corresponding night dataset generated for the aforementioned three rows.}
\label{fig:sf_xl_nc}
\end{figure*}

We trained NEG-CUT \cite{negcut} on the NightCities dataset and used it to process a subset of SF-XL \cite{cosplace}, resulting in SF-XL-NC. As shown in Fig. \ref{fig:sf_xl_nc}, we present some examples demonstrating our success in coupling nighttime knowledge with other desired knowledge for VPR.

\subsection{Datasets of VPR}

We test all models on a large number of night test sets, which helps to thoroughly understand each method’s strength and weaknesses. Below is a short description for each of the datasets.

\subsubsection{SF-XL Night} \cite{cosplace} is a huge dataset covering the whole city of San Francisco with over 41M images. Its test set covers the same with a less dense set of 2.8M images. SF-XL night queries \cite{reranking} are used which downloaded from Flickr for the area of San Francisco.

\subsubsection{Tokyo 24/7} \cite{netvlad} is a challenging dataset from the center of Tokyo. The database is made from GoogleStreetView, whereas the queries are a collection of smartphone photos from 105 places, and each place is photographed during the day, at sunset and at night. This results in 315 queries, each to be geolocalized independently.

\subsubsection{MSLS Night} \cite{msls} is the Mapillary Street-Level Sequences dataset, which has been created for image and sequence-based VPR. The dataset consists of more than 1M images from multiple cities, although only a small subset is used for evaluation.

\subsubsection{SVOX Night} \cite{sxov} is a cross-domain VPR dataset collected in a variety of weather and lighting conditions. It includes a large-scale database sourced from GoogleStreetView images spanning the city of Oxford. The night queries are extracted from the Oxford RobotCar dataset \cite{robotcar}.

\subsubsection{Aachen v1.1} \cite{vl_benchmarking} depicts the old inner city of Aachen, Germany. The database images used to build the reference scene representations were all taken during daytime with hand-held cameras over a period of about two years. The datasets offers query images taken at daytime and at nighttime. All query images were taken with mobile phone cameras, i.e., the Aachen Day-Night dataset considers the scenario of localization using mobile devices, e.g., for Augmented or Mixed Reality. The nighttime query images were taken using software HDR to create (relatively) high-quality images that are well-illuminated.

\subsubsection{RobotCar Seasons v2} \cite{vl_benchmarking} depicts the city of Oxford, UK. The reference and query images were captured by three synchronized cameras mounted on a car, pointing to the rear-left, rear, and rear-right, respectively. The images were recorded by driving the same route over a period of 12 months. One traversal is used to define a reference condition and the reference scene representation. Other traversals, covering different seasonal and illumination conditions, are used for query. All images were recorded in sequences. The RobotCar Seasons dataset represents an autonomous driving scenario, where it is necessary to localize images taken under varying seasonal conditions against a (possibly outdated) reference scene representation. Compared to the Aachen Day-Night dataset, the nighttime images of the RobotCar Seasons dataset exhibit significantly more motion blur and are of lower image quality.

%
%
\bibliographystyle{splncs04}
\bibliography{main}

\end{document}